\documentclass[11pt,a4paper]{article}
\usepackage[hyperref]{naaclhlt2019}
\usepackage{times}
\usepackage{latexsym}
\usepackage{url}

\aclfinalcopy 
\def\aclpaperid{1634} 

\title{Unsupervised Dialog Structure Learning
}

\author{Weiyan Shi \\
  University of California, Davis \\
  {\tt wyshi@ucdavis.edu} \\\And
  Tiancheng Zhao \\
  Carnegie Mellon University \\
  {\tt tianchez@cs.cmu.edu} \\\And
  Zhou Yu \\
  University of California, Davis \\
  {\tt joyu@ucdavis.edu}}

\date{}

\usepackage{amsmath,array,graphicx,amssymb}
\usepackage[export]{adjustbox}
\usepackage{caption}
\usepackage{subcaption}
\usepackage{multirow}
\usepackage{mathtools} 
\usepackage{graphicx} 
\usepackage{tabularx}
\usepackage{breakcites}
\usepackage{caption}
\usepackage{xcolor}
\usepackage{array}
\usepackage{makecell}
\usepackage{tabularx}
\usepackage{mathrsfs}
\usepackage{subcaption}
\usepackage{scalerel}

\usepackage{algorithm}
\usepackage{algorithmic}
\usepackage{mdframed}
\usepackage{tikz}
\usetikzlibrary{shapes,arrows,automata,positioning,cd, calc}
    \tikzset{
      cfgedge/.style   = {black, ->, >=stealth},
      forward/.style = { blue, ->, >=angle 45},
      backward/.style = { red, densely dashed, ->, >=latex' },
      backwardleft/.style = { red, densely dashed, <-, >=latex' },
    }
\usepackage{xcolor}
\usepackage{float}

\begin{document}
\maketitle
\begin{abstract}
Learning a shared dialog structure from a set of task-oriented dialogs is an important challenge in computational linguistics. The learned dialog structure can shed light on how to analyze human dialogs, and more importantly contribute to the design and evaluation of dialog systems. We propose to extract dialog structures using a modified VRNN model with discrete latent vectors. Different from existing HMM-based models, our model is based on variational-autoencoder (VAE). Such model is able to capture more dynamics in dialogs beyond the surface forms of the language. We find that qualitatively, our method extracts meaningful dialog structure, and quantitatively, outperforms previous models on the ability to predict unseen data. We further evaluate the model's effectiveness in a downstream task, the dialog system building task. Experiments show that, by integrating the learned dialog structure into the reward function design, the model converges faster and to a better outcome in a reinforcement learning setting.\footnote{The code is released at \url{https://github.com/wyshi/Unsupervised-Structure-Learning}}




\end{abstract}

\section{Introduction}
Human dialogs are like well-structured buildings, with words as the bricks, sentences as the floors, and topic transitions as the stairs connecting the whole building. Therefore, discovering dialog structure is crucial for various areas in computational linguistics, such as  dialog system building \cite{young2006using}, discourse analysis \cite{grosz1986attention}, and dialog summarization \cite{murray2005extractive, liu2010dialogue}. 
In domain specific tasks such as restaurant booking, it's  common for  people to follow a typical conversation flow. 
Current dialog systems require human experts to design the dialog structure, which is time consuming and sometimes insufficient to satisfy various customer needs.
Therefore, it's of great importance to automatically discover dialog structure from existing human-human conversations  and incorporate it into the dialog system design.

However, modeling human conversation is not easy for machines. Some previous work rely on  human annotations to learn dialog structures in supervised learning settings \cite{jurafsky1997switchboard}. But since human labeling is expensive and hard to obtain, such method is constrained by the small size of training
examples, and by the limited number of application domains \cite{zhai2014discovering}. Moreover, structure annotations on human conversation can be subjective, which makes it hard to reach inter-rater agreements. Therefore, we propose an unsupervised method to infer the latent dialog structure since unsupervised methods do not require annotated dialog corpus. 

Limited previous work has studied unsupervised methods to model the latent dialog structure. Most of the previous methods use the \textit{hidden Markov model} to capture the temporal dependency within human dialogs \cite{chotimongkol2008learning,ritter2010unsupervised, zhai2014discovering}. We propose to adopt a new type of models, the \textit{variational recurrent neural network} (VRNN, a recurrent version of the VAE) \cite{chung2015recurrent}, and infer the latent dialog structure with variational inference. 
VRNN is suitable for modeling sequential information. Compared to the simpler HMM models, VRNN also has the flexibility to model highly non-linear dynamics \cite{chung2015recurrent} in human dialogs. 



Our basic approach assumes that the dialog structure is composed of a sequence of latent states. 
Each conversational exchange (a pair of user and system utterances at time $t$) belongs to  a \textit{latent state}, which has causal effect  on the future latent states and the words the conversants
produce. 
Because discrete latent states are more interpretable than continuous ones, we combine VRNN with Gumbel-Softmax \cite{jang2016categorical} to obtain discrete latent vectors to represent the latent states. A common way to represent the dialog structure both visually and numerically is  to construct a transition probability table among latent states. The idea of transition table inspires us to develop two model variants to model the dependency between states indirectly and directly.

Once we obtain such a human-readable dialog structure, we can use it to facilitate many downstream tasks, such as dialog system training. The motivation is that the dialog structure contains important information on the flow of the conversation; if the automatic dialog system can mimic the behaviour in human-human dialogs, it can interact with users in a more natural and user-friendly way. Therefore, we propose to integrate the dialog structure information into the reward design of the reinforcement learning (RL). 
Experiments show that the model with the proposed reward functions converges faster to a better success rate.

\section{Related Work}
Variational Autoencoders (VAEs) \cite{kingma2013auto, doersch2016tutorial, kingma2014semi} have gained popularity in many computational linguistics tasks due to its interpretable  generative model structure \cite{miao2016language, miao2017discovering}. 
\citet{zhao2018unsupervised} applied VAE to learn discrete sentence representations and achieved good results. This is similar to our work, but we focus more on modeling the dialog structure and using the learned structure to improve the dialog system training.
\citet{serban2017hierarchical} presented a VHRED model which combines the VRNN and encoder-decoder structure for direct dialog response generation. While also similar, our model uses discrete latent vectors instead of continuous ones, and  re-constructs the utterances  to recover the latent  dialog structure, instead of modeling the responses directly.

There are some previous studies on discovering latent structure of conversations \cite{chotimongkol2008learning, ritter2010unsupervised, zhai2014discovering}. 
But they are all based on Hidden Markov Model (HMM). \citet{ritter2010unsupervised} extended the HMM-based method in \citet{chotimongkol2008learning} by adding additional word sources to social interaction data on Twitter. 
\citet{zhai2014discovering} decoupled the number of topics and the number of states to allow an additional layer of information in task-oriented dialogs. Our work also focuses on task-oriented dialogs but adopts the VRNN to perform variational inference in the model. According to \citet{chung2015recurrent}, the VRNN retains the flexibility to model highly non-linear dynamics, compared to simpler Dynamic Bayesian Network models such as HMM.

\begin{figure*}[h]
\centering
        \begin{subfigure}[b]{0.4\linewidth}
        \centering
          \begin{tikzpicture}[auto,
          node distance=1.5cm,  
          ] 
          \tikzstyle{every node} = [circle, draw,minimum size=0.5cm]

          \node (1) {$z_t$};
          \node [draw, diamond, minimum size=0.3cm, below of=1, text centered, inner sep=1pt, text width=0.4cm] (2) {$h_t$};
          \node [below of=2] (3) {$x_t$};
          \node [draw, diamond, minimum size=0.3cm, left of=2, text centered, inner sep=1pt, text width=0.45cm] (4) {$h_{\text{t-1}}$};

          \node [left of=1, text centered, inner sep=1.8pt] (5) {$z_{\text{t-1}}$};

          \path (4) edge[cfgedge, red, solid, thick] (1);
          
          \path (1) edge[cfgedge, gray, dashdotted, very thick] (2);
          \path (3) edge[cfgedge, gray, dashdotted, very thick] (2);
          \path (4) edge[cfgedge, gray, dashdotted, very thick] (2);
          
          \path (4) edge[cfgedge, blue, very thick, dotted] (3); 
          \path (1) edge [cfgedge,bend left=57, blue, very thick, dotted] (3);

          \path (4) edge[cfgedge,bend left=40, dashed, thick] (1); 
          \path (3) edge [cfgedge,bend left=47, dashed, thick] (1);

          \end{tikzpicture}
            \caption{Discrete-VRNN}
            \label{fig:model,prior1}
          \end{subfigure}
        \qquad
        \begin{subfigure}[b]{0.4\linewidth}
        \centering
          \begin{tikzpicture}[auto,
          node distance=1.5cm,  
          ] 
          \tikzstyle{every node} = [circle, draw,minimum size=0.5cm]

          \node (1) {$z_t$};
          \node [draw, diamond, minimum size=0.3cm, below of=1, text centered, inner sep=1pt, text width=0.4cm] (2) {$h_t$};
          \node [below of=2] (3) {$x_t$};
          \node [draw, diamond, minimum size=0.3cm, left of=2, text centered, inner sep=1pt, text width=0.45cm] (4) {$h_{\text{t-1}}$};

          \node [left of=1, text centered, inner sep=1.8pt] (5) {$z_{\text{t-1}}$};

          \path (5) edge[cfgedge, red, solid, thick] (1);
          
          \path (1) edge[cfgedge, gray, dashdotted, very thick] (2);
          \path (3) edge[cfgedge, gray, dashdotted, very thick] (2);
          \path (4) edge[cfgedge, gray, dashdotted, very thick] (2);
          
          \path (4) edge[cfgedge, blue, very thick, dotted] (3); 
          \path (1) edge [cfgedge,bend left=57, blue, very thick, dotted] (3);

          \path (4) edge[cfgedge,bend left=40, dashed, thick] (1); 
          \path (3) edge [cfgedge,bend left=47, dashed,thick] (1);

          \end{tikzpicture}
            \caption{Direct-Discrete-VRNN}
            \label{fig:model,prior2}
          \end{subfigure}
        \qquad

\caption{Discrete-VRNN (D-VRNN) and Direct-Discrete-VRNN (DD-VRNN) overview. D-VRNN and DD-VRNN use different priors to model the transition between $\textbf{z}_\text{t}$, shown in red solid lines. The regeneration of $\textbf{x}_\text{t}$ is in blue dotted lines. The recurrence of the state-level RNN is in gray dash-dotted lines. The inference of $\textbf{z}_\text{t}$ is in black dashed lines.}
\label{fig:model}
\end{figure*}
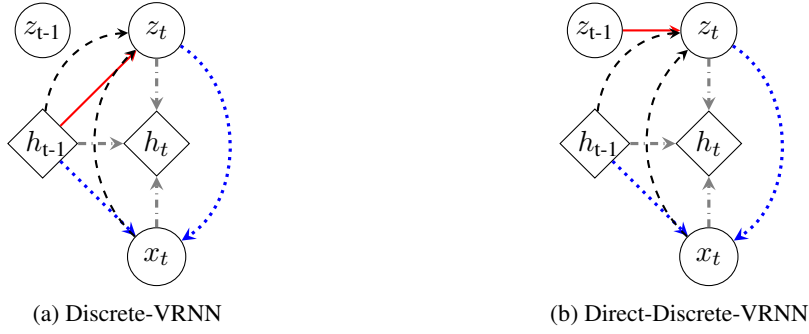

\citet{gunasekaraquantized} described a Quantized-Dialog Language Model (QDLM) for task-oriented dialog systems, which  performs clustering on utterances and models the dialog as a sequence of clusters to predict future responses. The idea of dialog discretization is similar to our method, but we choose VAE over simple clustering to allow more context-sensitivity and to capture more dynamics in the dialog beyond surface forms of the conversation. Additionally, we propose to utilize the dialog structure information to improve the dialog system training. 

Traditional reward functions in RL dialog training use delayed reward to provide feedback to the model. However, delayed reward suffers from potential slow convergence rate problem, so some studies integrated estimated per-turn immediate reward. For example, \citet{ferreira2013expert} studied expert-based reward shaping in dialog management. We use the KL-divergence between the transition probabilities and the predicted probabilities as the immediate per-turn reward. Different from the expert-based reward shaping, such reward does not require any manual labels and is generalizable to different tasks. 

%

\section{Models}

%
%
%

Fig.~\ref{fig:model} gives an overview of the Discrete-VRNN (D-VRNN) model and the Direct-Discrete-VRNN (DD-VRNN) model.
In principal, the VRNN contains a VAE at every timestep, and these VAEs are connected by a state-level RNN. 
The hidden state variable $\textbf{h}_{\text{t-1}}$ in this RNN encodes the dialog context up to time $t$. 
This connection helps the VRNN to model the temporal structure of the dialog \cite{chung2015recurrent}. 
The observed inputs $\textbf{x}_t$ to the model is the constructed utterance embeddings. $\textbf{z}_t$ is the latent vector in the VRNN at time $t$. Different from \citet{chung2015recurrent}, $\textbf{z}_t$ in our model is a discrete one-hot vector of dimension $N$, where $N$ is the total number of latent states. 

The major difference between D-VRNN and DD-VRNN lies in the priors of $\textbf{z}_t$.
In D-VRNN, we assume that $\textbf{z}_t$  depends on the entire dialog context $\textbf{h}_{\text{t-1}}$, shown in red in Fig.~1(a), which is the same as in \citet{chung2015recurrent}; while in DD-VRNN we assume that in the prior, $\textbf{z}_t$ directly depends on $\textbf{z}_{\text{t-1}}$ in order to model the direct transition between different latent states, shown in red in Fig.~1(b). 
We use $\textbf{z}_t$ and $\textbf{h}_{\text{t-1}}$ to regenerate the current utterances $\textbf{x}_t$ instead of generating the next utterances $\textbf{x}_{\text{t+1}}$, shown in blue dotted lines in Fig.~\ref{fig:model}. The idea of regeneration helps recover the dialog structure. 
Next, the recurrence in the RNN takes $\textbf{h}_{\text{t-1}}$, $\textbf{x}_t$ and $\textbf{z}_t$ to update itself, and allows the context to be passed down as the dialog proceeds, shown in gray dash-dotted lines.  
Finally in the inference, we construct the posterior of $\textbf{z}_t$ with the context $\textbf{h}_{t-1}$ and  $\textbf{x}_t$, and infer $\textbf{z}_t$ by sampling from the posterior, shown in black dashed lines in Fig.~\ref{fig:model}.
The mathematical details of each operation are described below.  $\varphi_{\tau}^{(\cdot)}$  are highly flexible feature extraction functions such as neural networks. $\varphi_{\tau}^{\textbf{x}}$, $\varphi_{\tau}^{\textbf{z}}$, $\varphi_{\tau}^{\text{prior}}$, $\varphi_{\tau}^{\text{enc}}$, $\varphi_{\tau}^{\text{dec}}$ are feature extraction networks for the input $\textbf{x}$, the latent vector $\textbf{z}$, the prior, the encoder and the decoder.



\noindent \textbf{Sentence Embedding}. $u_t=[\text{w}_{1, \text{t}}, \text{w}_{2,\text{t}}, ... {\text{w}_{\text{n}_{\text{w}}, t}}]$ and $s_t=[\text{v}_{1, t}, \text{v}_{2,t}, ... {\text{v}_{\text{n}_{\text{v}}, t}}]$ are the user utterance and the system utterance at time $t$, where $\text{w}_{\text{i}, \text{j}}$ and $\text{v}_{\text{i}, \text{j}}$ are individual words. The concatenation of  utterances from both parties, $x_t = [u_t, s_t]$, is the observed variable in the VAE. 
We use \citet{mikolov2013distributed} to perform word embedding and the average of the word embedding vectors of $u_t$ and $s_t$ are $\overline{u_t}$ and $\overline{s_t}$. The concatenation of $\overline{u_t}$ and $\overline{s_t}$ is used as the feature extraction of $\textbf{x}_t$, namely $    \varphi_{\tau}^{\textbf{x}}(\textbf{x}_t) = [\overline{u_t}, \overline{s_t}]
$. $\varphi_{\tau}^{\textbf{x}}(\textbf{x}_t)$ is the model inputs.


\noindent\textbf{Prior in D-VRNN}. The prior quantifies our assumption on $\textbf{z}_t$ before we observe the data. In the D-VRNN, it's reasonable to assume that the prior of $\textbf{z}_t$ depends on the context $\textbf{h}_{\text{t-1}}$ and follows the distribution shown in Eq.~(\ref{equ:priorequation}), because
conversation context is a critical factor that influences dialog transitions. 
Since $\textbf{z}_t$ is discrete, we use softmax to obtain the distribution. 

\begin{equation}
\label{equ:priorequation}
    \textbf{z}_t \sim \text{softmax}(\varphi_{\tau}^{\text{prior}}(\textbf{h}_{t-1}))
\end{equation}

\noindent\textbf{Prior in DD-VRNN}.
The dependency of $\textbf{z}_\text{t}$ on the entire context $\textbf{h}_{\text{t-1}}$ in Eq. (\ref{equ:priorequation}) makes it difficult to disentangle the relation between $\textbf{z}_\text{t-1}$ and $\textbf{z}_\text{t}$. But this relation is crucial in decoding how conversations flow from one conversational exchange to the next one. So in DD-VRNN, we directly model the influence of $\textbf{z}_\text{t-1}$ on $\textbf{z}_\text{t}$ in the prior, shown in Eq.~(\ref{equ:priorequation2}) and Fig.~1(b). To fit this prior distribution into the variational inference framework, we approximate $p(\textbf{z}_\text{t}|\textbf{x}_{<t}, \textbf{z}_{<t})$ with $p(\textbf{z}_\text{t}| \textbf{z}_{\text{t-1}})$ in Eq.~(\ref{equ:approximate}). Later, we show that the designed new prior has benefits under certain scenarios.

\begin{equation}
\label{equ:priorequation2}
    \textbf{z}_t \sim \text{softmax}(\varphi_{\tau}^{\text{prior}}(\textbf{z}_{\text{t-1}}))
\end{equation}
\vspace{-0.8cm}
\begin{equation}
\label{equ:approximate}
\begin{split}
p(\textbf{x}_{\leq T}, \textbf{z}_{\leq T}) &= \prod^T_{t=1} p(\textbf{x}_t | \textbf{z}_{\leq t}, \textbf{x}_{\leq t})p(\textbf{z}_t|\textbf{x}_{<t}, \textbf{z}_{<t})\\
&\approx \prod^T_{t=1} p(\textbf{x}_t | \textbf{z}_{\leq t}, \textbf{x}_{\leq t})p(\textbf{z}_t| \textbf{z}_{\text{t-1}})\\
\end{split}
\end{equation}

\noindent \textbf{Generation}.
$\textbf{z}_t$ is a summarization of the current conversational exchange under the context. We use $\textbf{z}_t$ and $\textbf{h}_{\text{t-1}}$ to reconstruct the current utterance $\textbf{x}_t$. This regeneration of $\textbf{x}_t$ allows us to recover the dialog structure.

We use two RNN decoders, \textit{dec1} and \textit{dec2}, parameterized by $\gamma_1$ and $\gamma_2$ to generate the original $u_t$ and $s_t$ respectively. $c_t$ and $d_t$ are the hidden states  of \textit{dec1} and \textit{dec2}. The context $\textbf{h}_{\text{t-1}}$ and feature extraction vector $\varphi_{\tau}^{\textbf{z}}(\textbf{z}_t)$ are concatenated to form the initial hidden state $h_0^{\text{dec1}}$ of \textit{dec1}. $c_{(n_w, t)}$ is the last hidden state of \textit{dec1}. Since $v_t$ is the response of $u_t$ and will be affected by $u_t$, we 
concatenate $c_{(n_w, t)}$ to $d_0$ to pass the information from $u_t$ to $v_t$. This concatenated vector is used as $h_0^{\text{dec2}}$ of \textit{dec2}. This process is shown in Eq.~(\ref{equ:dec1}) and (\ref{equ:dec2}). 
\vspace{-0.3cm}
\begin{equation}
\label{equ:dec1}
\begin{split}
&c_0 = [\textbf{h}_{t-1}, \varphi_{\tau}^{\textbf{z}}(\textbf{z}_t)],\\
&w_{(i, t)}, c_{(i, t)} = f_{\gamma_1}( w_{(i-1, t)},  c_{(i-1, t)})
\end{split}
\end{equation}
\vspace{-0.3cm}
\begin{equation}
\label{equ:dec2}
\begin{split}
&d_0 = [\textbf{h}_{t-1}, \varphi_{\tau}^{\textbf{z}}(\textbf{z}_t), c_{(n_w, t)}],\\
&v_{(i, t)}, d_{(i, t)} = f_{\gamma_2}( v_{(i-1, t)},  d_{(i-1, t)})
\end{split}
\end{equation}

\noindent \textbf{Recurrence}. 
The state-level RNN updates its hidden state $\textbf{h}_t$ with  $\textbf{h}_{\text{t-1}}$ based on the following Eq.~(\ref{recurrence_equation}). $f_{\theta}$ is a RNN parameterized by $\theta$. 
\vspace{-0.2cm}
\begin{equation}
\label{recurrence_equation}
    \textbf{h}_t = f_\theta(\varphi_{\tau}^{\textbf{z}}(\textbf{z}_t), \varphi_{\tau}^{\textbf{x}}(\textbf{x}_t), \textbf{h}_{t-1}) 
\end{equation}

\noindent \textbf{Inference}. We  infer $\textbf{z}_t$ from  the context $\textbf{h}_{\text{t-1}}$ and current utterances $\textbf{x}_t$, and construct the posterior distribution of $\textbf{z}_t$  by another softmax, shown in Eq.~(\ref{equ:inference}). Once we have the posterior distribution, we apply Gumbel-Softmax to take samples of $\textbf{z}_t$. D-VRNN and DD-VRNN differ in their priors but not in their inference, because 
we assume the direct transitions between $\textbf{z}_t$ in the prior instead of in the inference.


\vspace{-0.2cm}
\begin{equation}
    \textbf{z}_t | \textbf{x}_t \sim \text{softmax}([\varphi_{\tau}^{\text{enc}}(\textbf{h}_{t-1}), \varphi_{\tau}^{\textbf{x}}(\textbf{x}_t)])
\label{equ:inference}
\end{equation}

\noindent \textbf{Loss function}.
The objective function of VRNN is a timestep-wise variational lower bound, shown in Eq.~(\ref{equ:VRNN loss}) \cite{chung2015recurrent}.
To mitigate the \textit{vanishing latent variable problem} in VAE, we incorporate bow-loss and Batch Prior Regularization (BPR) \cite{zhao2017learning,zhao2018unsupervised} with tunable weights, $\lambda$ to the final loss function, shown in Eq.~(\ref{equ:VRNN loss bow}).

\vspace{-0.3cm}
\begin{equation}
\label{equ:VRNN loss}
\begin{split}
 &\mathcal{L}_{\text{VRNN}}= \mathbb{E}_{q(\textbf{z}_{\leq T}|\textbf{x}_{\leq T})}[ \log p(\textbf{x}_t \mid \textbf{z}_{\leq t}, \textbf{x}_{<t})) + \\
 &\sum\limits_{t=1}^{T}\text{-KL}(q(\textbf{z}_t \mid \textbf{x}_{\leq t}, \textbf{z}_{<t})\parallel 
 p(\textbf{z}_t \mid \textbf{x}_{< t}, \textbf{z}_{<t}))]
\end{split}
\end{equation}
\vspace{-0.8cm}
\begin{equation}
\label{equ:VRNN loss bow}
    \;\;
    \;\;\mathcal{L}_{\text{D-VRNN}} = \mathcal{L}_{\text{VRNN-BPR}} + \lambda*\mathcal{L}_{\text{bow}}\;\;\;\;\;\;\;\;\;\;\;\;\;\;\;\;\;\;
\end{equation}

\subsection{Transition Probability Calculation}
\label{sec:transition probability}
A good way to represent a dialog structure both numerically and visually is to construct a transition probability table among latent states. Such transition probability can also be used to design reward function in the RL training process. We calculate transition table differently for D-VRNN and DD-VRNN due to their different priors.   

\noindent \textbf{D-VRNN}. From Eq.~(\ref{recurrence_equation}), we know that $\textbf{h}_t$ is a function of $\textbf{x}_{\leq t}$ and $\textbf{z}_{\leq t}$. Combining Eq.~(\ref{equ:priorequation}) and  (\ref{recurrence_equation}), we find that $\textbf{z}_t$ is a function of $\textbf{x}_{\leq t}$ and $\textbf{z}_{< t}$. Therefore, $\textbf{z}_{<t}$ has an indirect influence on $\textbf{z}_{t}$ through $\textbf{h}_{\text{t-1}}$. 
This indirect influence reinforces our assumption that the previous  states $\textbf{z}_{<t}$ impacts future state $\textbf{z}_t$, but also makes it hard to recover a clear structure and disentangle the direct impact of $\textbf{z}_{\text{t-1}}$ on $\textbf{z}_t$. 

In order to better visualize the dialog structure and compare with the HMM-based models, we quantify the impact of $\textbf{z}_{\text{t-1}}$ on $\textbf{z}_t$ by estimating a bi-gram transition probability table, where $p_{i, j}=\frac{\# (\text{state}_i, \text{state}_j)}{\# (\text{state}_i)}$. The numerator is the total number of the ordered tuples ($\text{state}_{\text{i, t-1}}$, $\text{state}_{\text{j, t}}$) and the denominator is the total number of $\text{state}_i$ in the dataset. We choose a bi-gram transition table over a n-gram transition table with a bigger $n$, as the most recent context is usually the most relevant, but it should be noted that unlike the HMM models, the degree of transition in our model is not limited nor pre-determined, because $\textbf{z}_t$ captures all the context. Depending on different applications, different $n$ may be selected. 

\noindent \textbf{DD-VRNN} As stated before, the dependency of $\textbf{z}_t$ on the entire context $\textbf{h}_{\text{t-1}}$ creates difficulty in calculating the transition table. This is our motivation to derive the prior in DD-VRNN. The outputs from the softmax  in the prior (Eq.~(\ref{equ:priorequation2})) directly constitute the transition table. So rather than estimating the transition probabilities by frequency count as in D-VRNN, we can optimize the loss function of DD-VRNN and  get the parameters in Eq.~(\ref{equ:priorequation2}) that directly form the transition table. 

\subsection{NE-D-VRNN}
In task-oriented dialog systems, the presence of certain named entities, such as food preference plays a crucial role in determining the phase of the dialog. To make sure the latent states capture such useful information, we assign larger weights on the named entities when calculating the loss function in Eq. (\ref{equ:VRNN loss bow}).  The weights encourage the reconstructed utterances to have more correct named entities, therefore influencing the latent state to have better representation. We refer this model as NE-D-VRNN (Named Entitiy Discrete-VRNN).

\begin{table*}[!htb]
\centering
    \begin{subtable}{.475\linewidth}
      \centering
        \begin{tabular}{llll} \hline
            From& \multicolumn{3}{l}{Utterance} \\ 
            \hline
            \hline
SYS:& \multicolumn{3}{p{0.8\linewidth}}{Okay, you don't care \textit{place}, do you? } \\
USR:& \multicolumn{3}{p{0.8\linewidth}}{That's correct.} \\
SYS:& \multicolumn{3}{p{0.8\linewidth}}{What date are you interested?} \\
USR:& \multicolumn{3}{p{0.8\linewidth}}{\textbf{Weather this morning.}} \\
SYS:& \multicolumn{3}{p{0.8\linewidth}}{\textbf{I believe you said this morning.}} \\

            \hline
\end{tabular}

\caption{One example in State 2, ``provide place and time''}

    \end{subtable}%
    \qquad
    \begin{subtable}{.475\linewidth}
      \centering
        \begin{tabular}{llll} \hline
            From& \multicolumn{3}{l}{Utterance} \\ 
            \hline
            \hline
USR:& \multicolumn{3}{p{0.8\linewidth}}{Weather tomorrow.} \\
SYS:& \multicolumn{3}{p{0.8\linewidth}}{[api\_call]. your weather report is here [report]. what else can I do?} \\
USR:& \multicolumn{3}{p{0.8\linewidth}}{\textbf{Weather this morning.}} \\
SYS:& \multicolumn{3}{p{0.8\linewidth}}{\textbf{I believe you said this morning.}} \\

            \hline
\end{tabular}
        
        \caption{One example in State 3, ``additional time request'' }

    \end{subtable} 
    \caption{The utterances in bold in both tables have similar surface forms but different context. They  are grouped in different latent states using the Discrete-VRNN. In contrast, they are  in the same cluster using the K-means.}
\label{table:context-sensitive clustering}
\end{table*}
\section{Datasets}
\label{sec:dataset}
We test the proposed method on the CamRest676 corpus, which was released and collected by \citet{wen2016network}. The task is to help users find restaurants in Cambridge, UK. While this task is highly similar to DSTC2, we choose this dataset instead of DSTC2 because it is relatively clean and comes with good entity extraction methods. There are a total of 676 dialogs in this dataset with three information slots (\textit{food, price range and area}) and three request table slots (\textit{address, phone and postcode}).

We also evaluate our model on another dataset of simulated conversations, proposed in \citet{zhao2018zero}. The task is to help users get the weather report in a certain place at a specific time. The dialog system is controlled by a fixed structure and hand-set probabilities. Therefore, learning the dialog structure of this dataset might be easier. 

We assume each latent vector in the VAE emits one conversational exchange, including one user utterance and the corresponding system response at time $t$, and each conversational exchange corresponds to one latent vector, following \citet{zhai2014discovering}.

\section{Experiments}

We use LSTM \cite{hochreiter1997long} with 200-400 units for the RNNs, and a fully-connected network for all the $\varphi_{\tau}^{(\cdot)}$ with a dropout rate of 0.4. Additionally, we use trainable 300-dimension  word embeddings initialized by Google word2vec \cite{mikolov2013distributed}. The maximum utterance word length is  40 and the maximum dialog length is 10. We set the $\lambda$ for the bow-loss  to be 0.1. 80\% of the entire dataset are used for training, 10\% for validation and  10\%  for testing. Parameters mentioned are selected based on the performance of the validation set. 

The evaluation of unsupervised methods has always been a challenge. We first compare our models with a simple K-means clustering algorithm to show its context sensitivity. Then we compare our models with traditional HMM methods both qualitatively and quantitatively. Finally, we compare the three proposed model variants. The qualitative evaluation involves generating dialog structures with different models, and the quantitative evaluations involves calculating the likelihood on a held-out test set under a specific model, which measures the model's predictive power.


\begin{figure*}[h]
\centering
\begin{minipage}[b]{.45\textwidth}
  \centering
  \includegraphics[width=0.9\columnwidth, height=5.5cm]{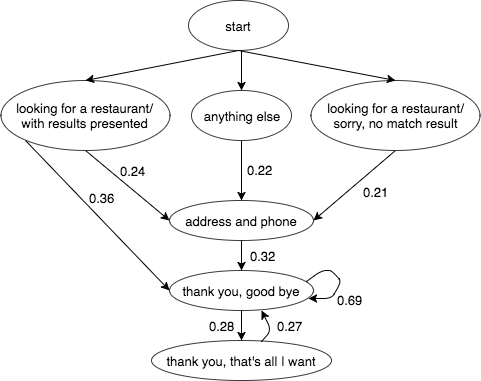}
\caption{Dialog structure of restaurant data by D-VRNN. Transitions with $P \geq 0.2$ are visualized. }
\label{fig:Structure for restaurant search data}
\end{minipage}\qquad
\begin{minipage}[b]{.45\textwidth}
  \centering
  \includegraphics[width=0.8\columnwidth, height=5.5cm]{graphs/vrnn/weather_vrnn.png}

\caption{Dialog structure of weather  data by D-VRNN. Transitions with $P \geq 0.2$ are visualized. }
\label{fig:Structure for simulated weather report data}
\end{minipage}
\end{figure*}

\subsection{Comparison with K-means Clustering}

We apply the model and obtain a latent state $\textbf{z}_t$ for each conversational exchange. Since $\textbf{z}_t$ is a discrete one-hot vector, we can group the conversational exchanges with the same latent state together. This process is similar to clustering the  conversational exchanges. But we choose the VAE over simple clustering methods because the VAE introduces more flexible context-sensitive information. A straightforward clustering method like K-means usually groups sentences with similar surface forms together, unless the previous context is explicitly encoded along with the  utterances. To compare the grouping result of our model with a traditional clustering method, we perform K-means clustering on  the dataset and calculate the within-cluster cosine similarity between the bag-of-word vectors of the utterances and the context. This cosine similarity measures how similar the utterances are on the word token level,  higher the value is, more words the utterances share in common. It turns out the average cosine similarity between the current utterance is 0.536 using the K-means and 0.357 using the D-VRNN, while the average cosine similarity between the context is 0.320 using the K-means and 0.351 using the D-VRNN. This does show that in the D-VRNN result, the context are more similar to each other, while in the K-means, the current utterances  are more similar on the surface textual level,
 which is not ideal because the dialog system needs context information. 
Table~\ref{table:context-sensitive clustering} shows an example where the D-VRNN clustering result is different from the K-means result. The conversational exchanges to be clustered are in the last row. These two exchanges have the same surface form but different contexts. D-VRNN identifies them as different by incorporating context information, whereas K-means  places them into the same cluster  ignoring the context.  

\subsection{Comparison with HMM}
HMM is similar to our model. 
Actually, if we remove the $\textbf{h}_{\text{t}}$ layer from Fig.~\ref{fig:model}, it becomes an HMM. But it is this additional layer of  $\textbf{h}_{\text{t}}$ that  encodes the dialog context into continuous information and is crucial in the success of our model.

\begin{figure}[h!]
\centering
  \includegraphics[width=0.95\columnwidth, height=5.5cm]{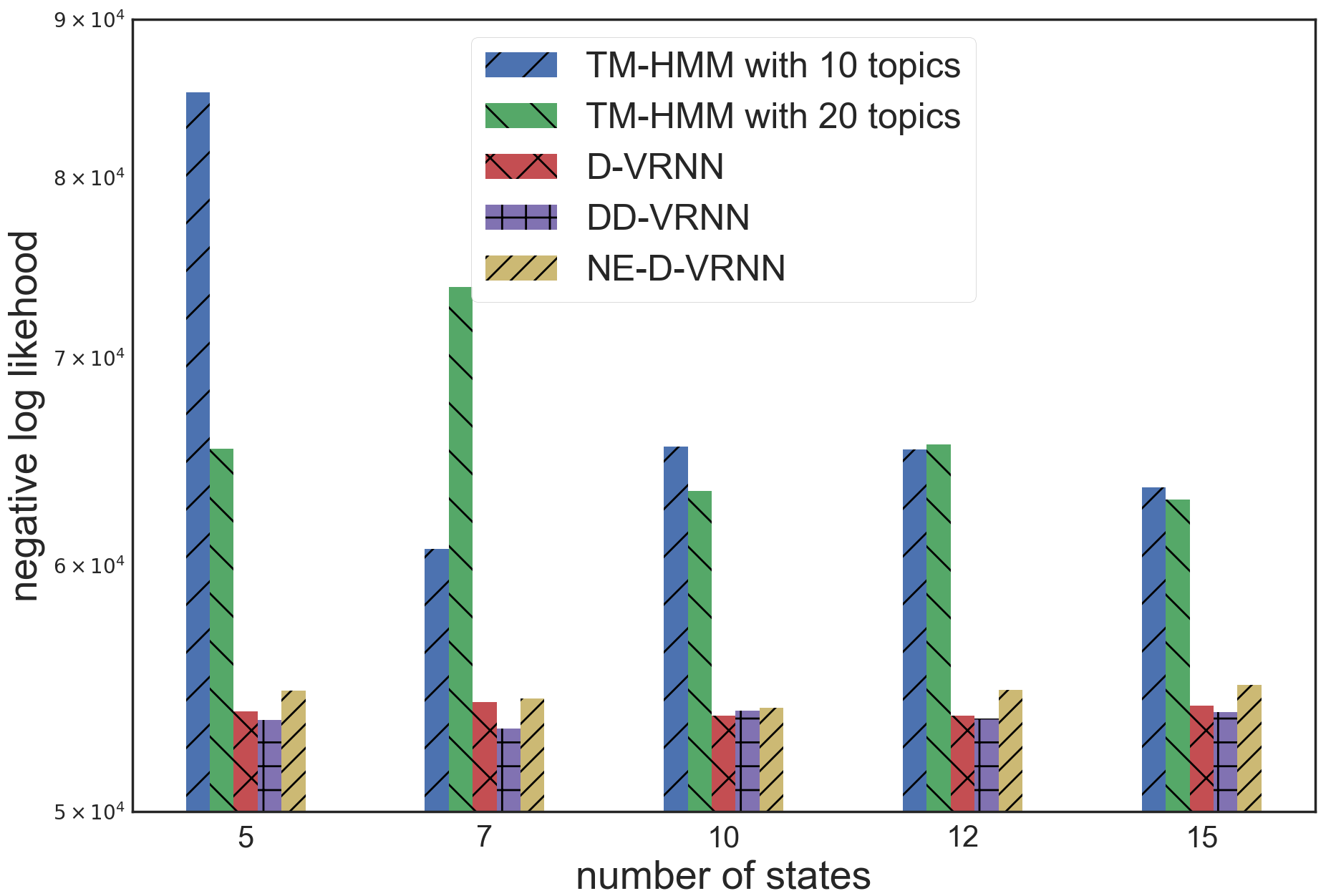}
  \caption{Negative log likelihood on restaurant data.}
  \label{fig:neg log likelihood1}
\end{figure}
\begin{figure}[h!]
\centering
  \includegraphics[width=0.95\columnwidth, height=5.5cm]{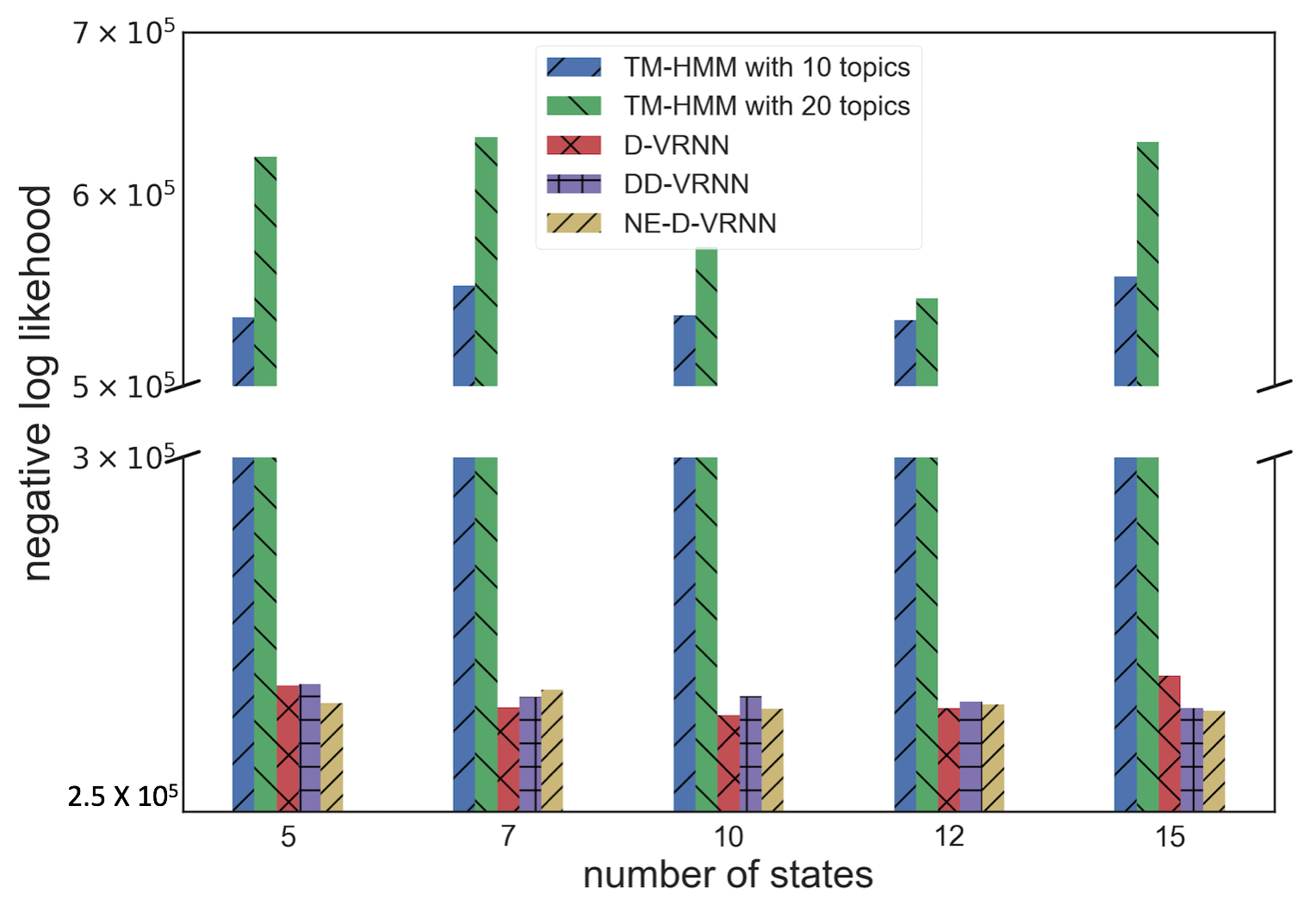}
  \caption{Negative log likelihood on  weather data.}
  \label{fig:neg log likelihood2}
\end{figure}

In Fig.~\ref{fig:neg log likelihood1} and ~\ref{fig:neg log likelihood2}, we compare our models quantitatively with the TM-HMM model with 10 topics and 20 topics from \citet{zhai2014discovering}, which performs the best on a similar task-oriented dialog dataset, the DSTC1 
.  
The y-axis  shows the negative log likelihood of reconstructing the test set under a specific model. 
The lower the negative log likelihood, the better the model performs.  The x-axis shows  different numbers of latent states $N$ used in the models. 
As we can see, all of the VRNN-based models surpass the TM-HMM models by a large margin and are more invariant to the change in $N$ on both datasets. Especially when $N$ is small, the performance of HMM is not stable.

Qualitatively, we compare the dialog structures generated by different models. Fig.~\ref{fig:Structure for restaurant search data} and \ref{fig:Structure for simulated weather report data} show the discovered dialog structures using the D-VRNN model, and Fig.~\ref{fig:dialog structure HMM} in the Appendix shows the dialog structures learned by HMM. Each circle in these figures represents a latent state $\textbf{z}_t$ with expert interpreted meanings, and the directed-arrows between the circles represent the transition probabilities between states. Human experts interpret each $\textbf{z}_t$ consistently by going through conversational exchanges assigned to the same $\textbf{z}_t$. 
 For a better visualization effect, we only visualize the transitions with a probability equal or greater than 0.2 in the figures. 

We observe reasonable dialog structures in Fig.~\ref{fig:Structure for restaurant search data} and \ref{fig:Structure for simulated weather report data}. The D-VRNN captures the major path \textit{looking for restaurant (anything else) $\rightarrow$ get restaurant address and phone $\rightarrow$ thank you} in the restaurant search task. It also captures \textit{what's the weather $\rightarrow$ place and time $\rightarrow$ api call} in the weather report task.  However, we do not get a dialog structure with entities separated (such as \textit{food type I like $\rightarrow$ area I prefer $\rightarrow$ price range I want $\rightarrow$ ...}). Because users know the system's capability and tend to give as many entities as possible in a single utterance,  so these entities are all mingled in one dialog exchange. But the model is able to distinguish  the ``presenting match result'' state from the ``presenting no match result'' state (on the top of Fig.~\ref{fig:Structure for restaurant search data}), which is important in making correct predictions. But in Fig.~7(a) generated by HMM, even if we set 10 states in the HMM, some states are still collapsed by the model because they share a similar surface form. And in Fig.~7(b) also by HMM, 
the dialog  flow skips the ``what can I do'' state, and goes from the ``start'' directly to ``providing place and time'', which is not reasonable.  

Another interesting phenomenon is that there are two ``thank you concentrated'' states in Fig.~\ref{fig:Structure for restaurant search data}. This is because, users frequently say ``thank you'' on two occasions,  1) after the dialog system presents the restaurant information, most users will say ``thank you''; 2) then the system will ask ``is there anything else I can help you with'', after which users typically respond with ``thank you, that's all''. This interesting structure is a reflection of the original rules in the dialog system. Moreover, in Fig.~\ref{fig:Structure for simulated weather report data}, we see 1) transitions from both directions between states ``place and time'' and ``api call'', and 2) transitions from ``api call'' to itself, 
 as there are simulated speech recognition errors in the data and the system needs to confirm the entity values and update the API query.



\subsection{VRNN Model Variants Comparison}
Even though the three proposed VRNN-based models have similar structures, they perform differently and are able to compensate each other. For example, 
in Fig.~8(b) in the Appendix, DD-VRNN is able to recognize a new state ``not done yet, what's the weather'', when users start a new query. This can complement D-VRNN's result. 

Quantitatively, the three model variants also perform differently. On the restaurant test set shown in Fig.~\ref{fig:neg log likelihood1}, DD-VRNN has the best overall performance compared with other models. Especially  when the number of states $N$ is small (e.g. 5 or 7 states), the advantage of the direct transition is more obvious. 
We think the reason behind is that it's easier and more accurate to model the direct transitions between a smaller set of states; as we increase $N$, the direct transitions between states become less and less obvious and therefore, help less on the predictive power. To our surprise, putting more weights on the named entities has a negative impact on the performance on the restaurant dataset. The underlying reason  might be that the emphasis on the named entities shifts the focus of the model from abstract latent representation to a more concrete word token level.  However, on the simulated weather dataset shown in Fig.~\ref{fig:neg log likelihood2}, NE-D-VRNN performs relatively well. It might be because the weather dataset is completely simulated, which makes it easier to recognize the named entities. With a more accurate named entity recognition, NE-D-VRNN is able to help the performance. Overall, D-VRNN is the most stable one across datasets, so we will use D-VRNN in the following experiments.

\section{Application on RL}
The ultimate goal of such a structure discovery model is to utilize the structure to facilitate downstream tasks, such as dialog system training. Therefore, 
we propose to incorporate the dialog structure information into the reward function of the RL  training. 
We believe that the transition table learned from the original dataset will guide the policy to make better decisions and converge faster by encouraging the policy to follow the real-data distribution. Similar to other RL  models, we build a user simulator to perform the RL training. Please refer to the Appendix for the training details. 

\subsection{Reward Function Design}
We use \textit{policy gradient} method \cite{williams1992simple} to train the dialog policy. 
Traditional reward functions give a positive reward (e.g. 20) after the successful completion of the task, 0 or a negative reward to penalize a failed task and -1 at each extra turn to encourage the system to finish the task sooner rather than later \cite{williams2017hybrid}. But this type of delayed reward functions doesn't have immediate rewards at each turn, which makes the model converge slowly. Therefore, we propose to incorporate the learned conversational exchange transition information as an immediate reward. The intuition is that in order to complete a task sooner, most users will follow a certain pattern when interacting with the system,  for example, a typical flow is that users first  give the entities information such as location, then ask for the restaurant information and finally, end the conversation.  If we can provide the RL model with the information on what action is more likely to follow  another action, the model can learn to follow real-data distributions and make better predictions. 
We encode the transition information through KL-divergence, a measurement of the distance between two distributions, in the reward function. 

We design four types of reward functions and describe each of them in Algorithm~\ref{alg:reward} and Eq.~\ref{equ:reward} in the Appendix. The traditional delayed reward  is the baseline. The second reward function, Rep-reward, uses constant penalty for repeated questions, as penalizing repetition  yields better results, according to \citet{shi2018sentiment}. 
The third reward function, KL-reward, incorporates the transition table information. From the RL model, we get the predicted probability $p_{\text{pred}}$ for different actions; from the D-VRNN model, we get the transition probability $p_{\text{trans}}$ between states and each state is translated to an action. We calculate the negation of the KL-divergence between   $p_{\text{trans}}$ and  $p_{\text{pred}}$ and use it as the immediate reward for every turn. This immediate reward links the predicted distribution with the real-data distribution by calculating the distance between them.  The fourth reward function (KL+Rep) gives an additional -2 repetition penalty to the KL-reward to test the combination effect of the two types of penalties.


\begin{figure}
\centering
  \includegraphics[width=1.1\columnwidth, height=6cm]{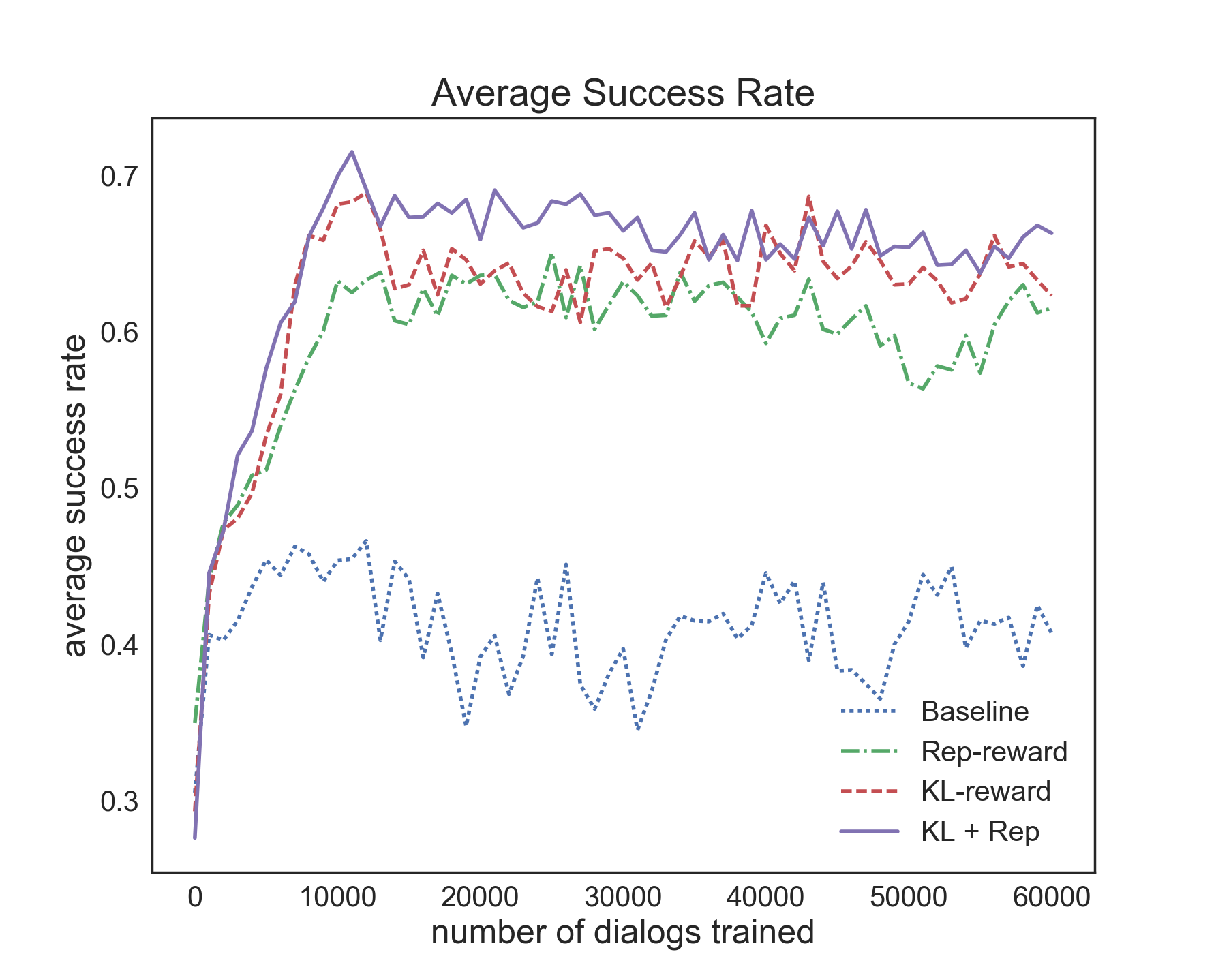}
  \caption{Average success rate of RL models with different reward functions.}
  \label{fig:RL_success}
\end{figure}

\subsection{Result Analysis}

We evaluate the RL performance by the average success rate, shown in Fig.~\ref{fig:RL_success}.
We observe  that all the experimental reward functions greatly improve the performance of the baseline. We also observe that the baseline has a higher variance, which is due to the inefficient delayed rewards. Moreover, both the KL-reward and the KL-Rep reward reach a higher success rate at around 10,000 iterations than the Rep-reward and converges faster. These two reward functions also achieve a better convergent success rate after 10,000 iterations. This suggests that adding KL-divergence of $p_{\text{trans}}$ and $p_{\text{pred}}$ into the reward helps develop a better policy faster. 

The good performance comes from the transition probability $p_{\text{trans}}$ learned by the discrete-VRNN. $p_{\text{trans}}$ summarizes the communication pattern of most users in the real world and the KL-divergence measures the distance between the predicted distribution  $p_{\text{pred}}$ and  $p_{\text{trans}}$ from the real dataset.
The RL model processes this KL-reward signal and learns to minimize the gap between $p_{\text{trans}}$ and $p_{\text{pred}}$. As a result, $p_{\text{pred}}$ will follow closer to the real data distribution, leading the model to converge faster to a better task success rate. In this way, we successfully incorporate the dialog structure information from the real data into the RL system training.

We observe that the use of KL reward improves the performance significantly in terms of both convergence rate and the final task success rate. Further, the combination of KL and repetition reward makes the model more stable and achieves a  better task success rate, compared with the model with  only KL-reward or Rep-reward. 
This indicates that the KL-reward can be combined with other type of rewards to achieve a better performance. 


\section{Conclusion and Future Work}
A key challenge for discourse analysis and dialog system building is to extract the latent dialog structure. We adopted the VRNN with discrete latent variables to learn the latent states of each conversational exchange and the transitions between these states in an unsupervised fashion. We applied the algorithm on a restaurant search task and a simulated weather report task, and evaluated the model quantitatively and qualitatively. We also proposed a way to incorporate the learned dialog structure information into a downstream dialog system building task.  We involved the dialog structure in the RL reward design, which made the model converge faster to a better task success rate. 

The performance of the Discrete-VRNN model has a major impact on the performance of the policy training. We plan to further improve the dialog structure learning process. Currently, we try to capture the status of the named entities by increasing the weights on the  entities, which focus on the concrete word token level. In the future, we may use more sophisticated ways to encode the entity information into the latent states. 

\bibliography{naaclhlt2019}

\begin{thebibliography}{28}
\expandafter\ifx\csname natexlab\endcsname\relax\def\natexlab#1{#1}\fi

\bibitem[{Chotimongkol(2008)}]{chotimongkol2008learning}
Ananlada Chotimongkol. 2008.
\newblock \emph{Learning the structure of task-oriented conversations from the
  corpus of in-domain dialogs}.
\newblock Ph.D. thesis, Carnegie Mellon University, Language Technologies
  Institute, School of Computer Science.

\bibitem[{Chung et~al.(2015)Chung, Kastner, Dinh, Goel, Courville, and
  Bengio}]{chung2015recurrent}
Junyoung Chung, Kyle Kastner, Laurent Dinh, Kratarth Goel, Aaron~C Courville,
  and Yoshua Bengio. 2015.
\newblock A recurrent latent variable model for sequential data.
\newblock In \emph{Advances in neural information processing systems}, pages
  2980--2988.

\bibitem[{Doersch(2016)}]{doersch2016tutorial}
Carl Doersch. 2016.
\newblock Tutorial on variational autoencoders.
\newblock \emph{arXiv preprint arXiv:1606.05908}.

\bibitem[{Ferreira and Lef{\`e}vre(2013)}]{ferreira2013expert}
Emmanuel Ferreira and Fabrice Lef{\`e}vre. 2013.
\newblock Expert-based reward shaping and exploration scheme for boosting
  policy learning of dialogue management.
\newblock In \emph{Automatic Speech Recognition and Understanding (ASRU), 2013
  IEEE Workshop on}, pages 108--113. IEEE.

\bibitem[{Grosz and Sidner(1986)}]{grosz1986attention}
Barbara~J Grosz and Candace~L Sidner. 1986.
\newblock Attention, intentions, and the structure of discourse.
\newblock \emph{Computational linguistics}, 12(3):175--204.

\bibitem[{Gunasekara et~al.(2017)Gunasekara, Nahamoo, Polymenakos, Ganhotra,
  and Fadnis}]{gunasekaraquantized}
R~Chulaka Gunasekara, David Nahamoo, Lazaros~C Polymenakos, Jatin Ganhotra, and
  Kshitij~P Fadnis. 2017.
\newblock Quantized-dialog language model for goal-oriented conversational
  systems.
\newblock \emph{Dialog State Tracking Challenge 6, COLIPS}.

\bibitem[{Hochreiter and Schmidhuber(1997)}]{hochreiter1997long}
Sepp Hochreiter and J{\"u}rgen Schmidhuber. 1997.
\newblock Long short-term memory.
\newblock \emph{Neural computation}, 9(8):1735--1780.

\bibitem[{Jang et~al.(2016)Jang, Gu, and Poole}]{jang2016categorical}
Eric Jang, Shixiang Gu, and Ben Poole. 2016.
\newblock Categorical reparameterization with gumbel-softmax.
\newblock \emph{arXiv preprint arXiv:1611.01144}.

\bibitem[{Jurafsky(1997)}]{jurafsky1997switchboard}
Dan Jurafsky. 1997.
\newblock Switchboard swbd-damsl shallow-discourse-function annotation coders
  manual.
\newblock \emph{Institute of Cognitive Science Technical Report}.

\bibitem[{Kingma et~al.(2014)Kingma, Mohamed, Rezende, and
  Welling}]{kingma2014semi}
Diederik~P Kingma, Shakir Mohamed, Danilo~Jimenez Rezende, and Max Welling.
  2014.
\newblock Semi-supervised learning with deep generative models.
\newblock In \emph{Advances in Neural Information Processing Systems}, pages
  3581--3589.

\bibitem[{Kingma and Welling(2013)}]{kingma2013auto}
Diederik~P Kingma and Max Welling. 2013.
\newblock Auto-encoding variational bayes.
\newblock \emph{arXiv preprint arXiv:1312.6114}.

\bibitem[{Liu et~al.(2010)Liu, Seneff, and Zue}]{liu2010dialogue}
Jingjing Liu, Stephanie Seneff, and Victor Zue. 2010.
\newblock Dialogue-oriented review summary generation for spoken dialogue
  recommendation systems.
\newblock In \emph{Human Language Technologies: The 2010 Annual Conference of
  the North American Chapter of the Association for Computational Linguistics},
  pages 64--72. Association for Computational Linguistics.

\bibitem[{Miao and Blunsom(2016)}]{miao2016language}
Yishu Miao and Phil Blunsom. 2016.
\newblock Language as a latent variable: Discrete generative models for
  sentence compression.
\newblock \emph{arXiv preprint arXiv:1609.07317}.

\bibitem[{Miao et~al.(2017)Miao, Grefenstette, and
  Blunsom}]{miao2017discovering}
Yishu Miao, Edward Grefenstette, and Phil Blunsom. 2017.
\newblock Discovering discrete latent topics with neural variational inference.
\newblock \emph{arXiv preprint arXiv:1706.00359}.

\bibitem[{Mikolov et~al.(2013)Mikolov, Sutskever, Chen, Corrado, and
  Dean}]{mikolov2013distributed}
Tomas Mikolov, Ilya Sutskever, Kai Chen, Greg~S Corrado, and Jeff Dean. 2013.
\newblock Distributed representations of words and phrases and their
  compositionality.
\newblock In \emph{Advances in neural information processing systems}, pages
  3111--3119.

\bibitem[{Murray et~al.(2005)Murray, Renals, and
  Carletta}]{murray2005extractive}
Gabriel Murray, Steve Renals, and Jean Carletta. 2005.
\newblock Extractive summarization of meeting recordings.

\bibitem[{Ritter et~al.(2010)Ritter, Cherry, and
  Dolan}]{ritter2010unsupervised}
Alan Ritter, Colin Cherry, and Bill Dolan. 2010.
\newblock Unsupervised modeling of twitter conversations.
\newblock In \emph{Human Language Technologies: The 2010 Annual Conference of
  the North American Chapter of the Association for Computational Linguistics},
  pages 172--180. Association for Computational Linguistics.

\bibitem[{Serban et~al.(2017)Serban, Sordoni, Lowe, Charlin, Pineau, Courville,
  and Bengio}]{serban2017hierarchical}
Iulian~Vlad Serban, Alessandro Sordoni, Ryan Lowe, Laurent Charlin, Joelle
  Pineau, Aaron~C Courville, and Yoshua Bengio. 2017.
\newblock A hierarchical latent variable encoder-decoder model for generating
  dialogues.
\newblock In \emph{AAAI}, pages 3295--3301.

\bibitem[{Shi and Yu(2018)}]{shi2018sentiment}
Weiyan Shi and Zhou Yu. 2018.
\newblock Sentiment adaptive end-to-end dialog systems.
\newblock \emph{arXiv preprint arXiv:1804.10731}.

\bibitem[{Tokic(2010)}]{tokic2010adaptive}
Michel Tokic. 2010.
\newblock Adaptive $\varepsilon$-greedy exploration in reinforcement learning
  based on value differences.
\newblock In \emph{Annual Conference on Artificial Intelligence}, pages
  203--210. Springer.

\bibitem[{Wen et~al.(2016)Wen, Vandyke, Mrksic, Gasic, Rojas-Barahona, Su,
  Ultes, and Young}]{wen2016network}
Tsung-Hsien Wen, David Vandyke, Nikola Mrksic, Milica Gasic, Lina~M
  Rojas-Barahona, Pei-Hao Su, Stefan Ultes, and Steve Young. 2016.
\newblock A network-based end-to-end trainable task-oriented dialogue system.
\newblock \emph{arXiv preprint arXiv:1604.04562}.

\bibitem[{Williams et~al.(2017)Williams, Asadi, and Zweig}]{williams2017hybrid}
Jason~D Williams, Kavosh Asadi, and Geoffrey Zweig. 2017.
\newblock Hybrid code networks: practical and efficient end-to-end dialog
  control with supervised and reinforcement learning.
\newblock \emph{arXiv preprint arXiv:1702.03274}.

\bibitem[{Williams(1992)}]{williams1992simple}
Ronald~J Williams. 1992.
\newblock Simple statistical gradient-following algorithms for connectionist
  reinforcement learning.
\newblock \emph{Machine learning}, 8(3-4):229--256.

\bibitem[{Young(2006)}]{young2006using}
Steve Young. 2006.
\newblock Using pomdps for dialog management.
\newblock In \emph{Spoken Language Technology Workshop, 2006. IEEE}, pages
  8--13. IEEE.

\bibitem[{Zhai and Williams(2014)}]{zhai2014discovering}
Ke~Zhai and Jason~D Williams. 2014.
\newblock Discovering latent structure in task-oriented dialogues.
\newblock In \emph{Proceedings of the 52nd Annual Meeting of the Association
  for Computational Linguistics (Volume 1: Long Papers)}, volume~1, pages
  36--46.

\bibitem[{Zhao and Eskenazi(2018)}]{zhao2018zero}
Tiancheng Zhao and Maxine Eskenazi. 2018.
\newblock Zero-shot dialog generation with cross-domain latent actions.
\newblock \emph{arXiv preprint arXiv:1805.04803}.

\bibitem[{Zhao et~al.(2018)Zhao, Lee, and Eskenazi}]{zhao2018unsupervised}
Tiancheng Zhao, Kyusong Lee, and Maxine Eskenazi. 2018.
\newblock Unsupervised discrete sentence representation learning for
  interpretable neural dialog generation.
\newblock \emph{arXiv preprint arXiv:1804.08069}.

\bibitem[{Zhao et~al.(2017)Zhao, Zhao, and Eskenazi}]{zhao2017learning}
Tiancheng Zhao, Ran Zhao, and Maxine Eskenazi. 2017.
\newblock Learning discourse-level diversity for neural dialog models using
  conditional variational autoencoders.
\newblock \emph{arXiv preprint arXiv:1703.10960}.

\end{thebibliography}
\bibliographystyle{acl_natbib}

\appendix
\section{Appendices}
\label{sec:supplemental}



\subsection{Reward Functions in RL}
\label{sec:rl reward}
\begin{algorithm}
\caption{Reward function}
\begin{algorithmic} 
\IF{success}
	\STATE $R = 20 $
\ELSIF{failure}
	\STATE $R = -10 $
\ELSIF{repeated question}
    \STATE $R = f_{\text{reward}}(p_{\text{trans}}, p_{\text{pred}})$
\ELSIF{each proceeding turn}
    \STATE $R = -1$
\ENDIF
\end{algorithmic}\label{alg:reward}
\end{algorithm}
\vspace{-0.8cm}
\begin{equation}
\begin{split}
    f_{\text{reward}}&(p_{\text{trans}}, p_{\text{pred}}) = \\
    &\begin{cases}
    -1 & \text{Baseline}\\
    -5                       & \text{Rep-reward}\\
    -\text{KL}(p_{\text{trans}}, p_{\text{pred}}) & \text{KL-reward}\\
    -\text{KL}(p_{\text{trans}}, p_{\text{pred}})-2 & \text{KL+Rep}
    \end{cases}
\end{split}
\label{equ:reward}
\end{equation}

\subsection{RL Simulation Setting}

The main purpose of the user simulator is to provide named entities. Therefore, it is built by  sampling utterances from the real dataset according to the current status of named entities. Table~\ref{tab:truthTables} shows one simulated dialog in the reinforcement learning. 
We design seven system template actions in the simulated setting which fall under four categories, \textit{ask for entity}, \textit{present results}, \textit{give restaurant information} and \textit{closing statement}.
But there are $n$ latent states learned from the real data, and $n$ is not necessarily equal to four. To solve this mismatch in dimensions between $p_\text{pred}$ and $p_\text{tran}$, human experts further collapse similar latent states and reduce the number of latent states  to four, which represents the four main categories in the actions, \textit{ask for entity}, \textit{present results}, \textit{give restaurant information} and \textit{closing statement}. For example, the two ``thank you concentrated'' states in Fig.~\ref{fig:Structure for restaurant search data} will be combined to the \textit{closing statement} category. The probability of each category is  the sum of the original probabilities whose actions are now under the same category.
\begin{table}[h]
\begin{tabular}{llll} 
\hline
            From& \multicolumn{3}{l}{Utterance} \\ 
            \hline
            \hline
USR:& \multicolumn{3}{p{0.7\linewidth}}{Could you please help me find an [value\_price] restaurant in the [value\_area] [slot\_area]?} \\
SYS:& \multicolumn{3}{p{0.7\linewidth}}{Do you have a [slot\_food] preference?} \\
USR:& \multicolumn{3}{p{0.7\linewidth}}{Let's try [value\_food], please} \\
SYS:& \multicolumn{3}{p{0.7\linewidth}}{[value\_name] is a good restaurant matching your request. Is there anything else I can help you with?} \\
USR:& \multicolumn{3}{p{0.7\linewidth}}{What other options are there?} \\
SYS:& \multicolumn{3}{p{0.7\linewidth}}{[value\_name] is a good restaurant matching your request. Is there anything else I can help you with?} \\
USR:& \multicolumn{3}{p{0.7\linewidth}}{Great. what is their [slot\_phone]?} \\
SYS:& \multicolumn{3}{p{0.7\linewidth}}{Here is the info...}
\\
\hline
\end{tabular}
    \caption{A simulated dialog.}
    \label{tab:truthTables}   
\end{table}

\subsection{RL Training Details}
A simple action mask is used to prevent impossible actions, such as presenting results without making a query to the DB. The input to the model is a list of common contextual features, such as the presence of each entity and the last action taken.
The output of the model is the system action template.

\begin{figure*}[htb!]
\centering
\begin{subfigure}{0.9\columnwidth}
    \centering
    \includegraphics[width=0.9\columnwidth,
    height=5.5cm]{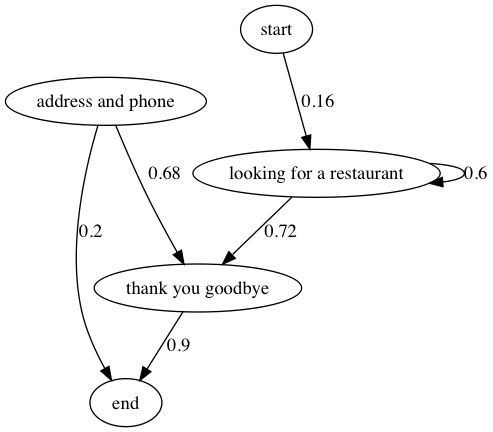}
    \caption{HMM, restaurant data, 5 states}
    \label{fig:hmm, restaurant}
\end{subfigure}
\qquad
\begin{subfigure}{0.9\columnwidth}
    \centering
    \includegraphics[width=0.9\columnwidth,
    height=5.5cm]{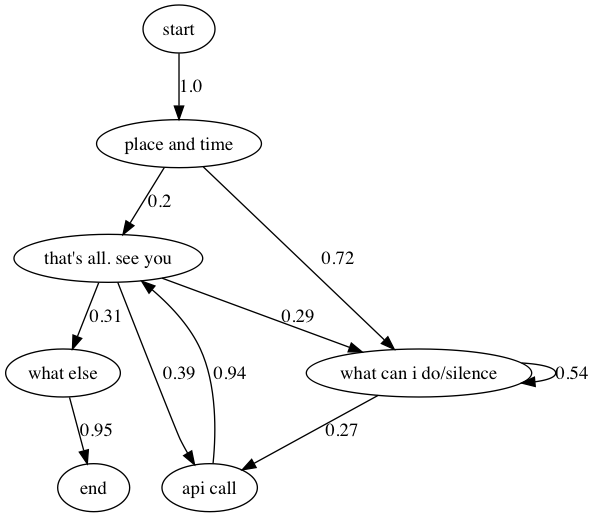}
    \caption{HMM, weather data, 10 states}
        \label{fig:hmm, weather}

\end{subfigure}


\caption{Dialog structures generated by HMM on different datasets. Transitions with $P \geq 0.2$ are visualized. }
\label{fig:dialog structure HMM}
\end{figure*}

\begin{figure*}[ht]
\centering
\begin{subfigure}{0.9\columnwidth}
    \centering
    \includegraphics[width=0.9\columnwidth, height=5.5cm]{graphs/ddvrnn/cambridge_ddvrnn.png}
    \caption{DD-VRNN, restaurant data, 5 states}
        \label{fig:ddvrnn, restaurant}

\end{subfigure}
\qquad
\begin{subfigure}{0.9\columnwidth}
    \centering
    \includegraphics[width=0.9\columnwidth, height=5.5cm]{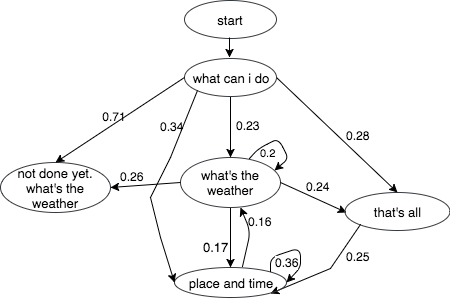}
    \caption{DD-VRNN, weather data, 10 states}
        \label{fig:ddvrnn, weather}

\end{subfigure}
\caption{Dialog structures generated by DD-VRNN on different datasets. Transitions with $P \geq 0.2$ are visualized. }
\label{fig:dialog structure DD-VRNN}
\end{figure*}

Four different reward functions are used. 
A discount factor of 0.9 is applied to all the experiments and the maximum number of turns is 10.
An LSTM with 32 hidden units is used and the RL policy is updated after each dialog. We also apply $\epsilon$-greedy exploration strategy \cite{tokic2010adaptive}. 
Because the RL training can sometimes be unstable, we initialize all the model parameters using supervised learning on a simulated dataset, which consists of 500 dialogs between the user simulator and a rule-based agent simulator.  
The method is evaluated by freezing the policy after every 1000 updates and running 200 simulated dialogs to calculate the task success rate. We repeat the entire process 10 times and report the average success rate in Fig.~\ref{fig:RL_success}.

\subsection{Dialog Structures by Different Models}
Fig.~\ref{fig:dialog structure HMM} and \ref{fig:dialog structure DD-VRNN} show the dialog structures generated by HMM and DD-VRNN.




\end{document}